\documentclass[pdflatex,sn-aps]{sn-jnl}

\usepackage{graphicx}%
\usepackage{multirow}%
\usepackage{amsmath,amssymb,amsfonts}%
\usepackage{amsthm}%
\usepackage{mathrsfs}%
\usepackage[title]{appendix}%
\usepackage{xcolor}%
\usepackage{textcomp}%
\usepackage{manyfoot}%
\usepackage{booktabs}%

\usepackage{makecell}
\usepackage{enumitem}
\usepackage{subcaption}
\usepackage[linesnumbered,algoruled,boxed,lined]{algorithm2e}
\DontPrintSemicolon
\usepackage{algorithmicx}%
\usepackage{listings}%
\usepackage{bm}
\usepackage{cleveref}
\Crefname{equation}{Eq.}{Eqs.}
\Crefname{figure}{Fig.}{Figs.}
\Crefname{tabular}{Tab.}{Tabs.}
\crefname{algocf}{alg.}{algs.}
\Crefname{algocf}{Algorithm}{Algorithms}
\Crefname{fct}{Fact}{Facts }

\theoremstyle{thmstyleone}%
\theoremstyle{thmstyletwo}%
\theoremstyle{thmstylethree}%

\newcommand{\ie}{i.e., }
\newcommand{\supth}[1]{\ensuremath{{#1}^{\text{th}}}}
\newcommand{\x}{\bm{x}}
\newcommand{\y}{\bm{y}}

\begin{document}

\title[Diffusion models for data assimilation]{Closed-form conditional diffusion models for data assimilation}


\author[1]{\fnm{Brianna} \sur{Binder}}\email{bjbinder@usc.edu}

\author[1,2]{\fnm{Agnimitra} \sur{Dasgupta}}\email{adasgu@sandia.gov}

\author*[1]{\fnm{Assad} \sur{Oberai}}\email{aoberai@usc.edu}

\affil*[1]{\orgdiv{Department of Aerospace \& Mechanical Engineering}, \orgname{University of Southern California}, \orgaddress{\street{3650 McClintock Ave}, \city{Los Angeles}, \postcode{90089}, \state{California}, \country{USA}}}

\affil[2]{\orgdiv{Optimization \& Uncertainty Quantification}, \orgname{Sandia National Laboratories}, \orgaddress{\street{1450 Innovation Pkwy SE}, \city{Albuquerque}, \postcode{87123}, \state{New Mexico}, \country{USA}}}


\abstract{We propose closed-form conditional diffusion models for data assimilation. Diffusion models use data to learn the score function (defined as the gradient of the log-probability density of a data distribution), allowing them to generate new samples from the data distribution by reversing a noise injection process. While it is common to train neural networks to approximate the score function, we leverage the analytical tractability of the score function to assimilate the states of a system with measurements. To enable the efficient evaluation of the score function, we use kernel density estimation to model the joint distribution of the states and their corresponding measurements. The proposed approach also inherits the capability of conditional diffusion models of operating in black-box settings, \ie the proposed data assimilation approach can accommodate systems and measurement processes without their explicit knowledge. The ability to accommodate black-box systems combined with the superior capabilities of diffusion models in approximating complex, non-Gaussian probability distributions means that the proposed approach offers advantages over many widely used filtering methods. We evaluate the proposed method on nonlinear data assimilation problems based on the Lorenz-63 and Lorenz-96 systems of moderate dimensionality and nonlinear measurement models. Results show the proposed approach outperforms the widely used ensemble Kalman and particle filters when small to moderate ensemble sizes are used.}




\maketitle

\section{Introduction}\label{sec:introduction}
Data assimilation (DA) involves estimating the state of a dynamical system from partial and noisy observations, and provides a fundamental framework for fusing observational data with numerical models~\cite{law2015data}. DA is ubiquitous across various fields of science and engineering, including the geosciences~\cite{carrassi2018data}, numerical weather prediction~\cite{navon2009data}, and structural engineering~\cite{impraimakis2022new}, among others. Broadly viewed, DA may be categorized as \emph{filtering} or \emph{smoothing}. Filtering involves sequentially updating the states of a dynamical system as new observations are made. Smoothing refers to inferring the entire trajectory of the dynamical system using all available observations. The focus of this paper is filtering, that is, we are interested in estimating the conditional distribution of the states of a dynamical system from their indirect, noisy and sparse measurements made sequentially in time. 

Bayesian filtering refers to the recursive application of Bayes' rule to estimate the conditional distribution or filter distribution of the states of a dynamical system at any instant of time, given the history of measurements made up to that point~\cite{sarkka2023bayesian}. Essentially, at every measurement instance, the filtering problem can be framed as a probabilistic inverse problem, which can be solved using Bayesian inference, and the corresponding posterior distribution is the filtering distribution. The Kalman filter provides an exact, closed-form solution to the posterior distribution for linear, Gaussian systems (\ie linear dynamical systems with Gaussian process noise and linear measurement models with Gaussian measurement noise)~\cite{kalman1960new}. However, the filtering distribution is generally intractable for nonlinear, non-Gaussian systems. Estimating the filtering distribution of such systems remains a long-standing computational challenge, and several advances have been made to approximate the filtering distribution in such cases. 

Nonlinear extensions to the Kalman filter, like the extended Kalman filter (EKF), unscented Kalman filter (UKF)~\cite{julier1997new}, and the ensemble Kalman filter (EnKF)~\cite{evensen2003ensemble,katzfuss2016understanding} all build Gaussian approximations to the filtering distribution. These approaches may be sufficient for near-linear systems with additive Gaussian process and measurement noise, but incur significant approximations for nonlinear systems with non-additive noise. To approximate non-Gaussian filtering distributions, particle filters, like the Sequential Importance Resampling (SIR) filter~\cite{doucet2000sequential,doucet2001introduction}, use a weighted set of particles to represent the filtering distribution. However, a common drawback of most particle filters is weight degeneracy: most particles attain near-zero weights after a few assimilation steps in high-dimensional settings~\cite{bickel2008sharp,snyder2008obstacles}. 

More recently, data assimilation strategies coupled with machine learning methods have been developed to improve the scalability and accuracy of the aforementioned traditional approaches. For example, attention maps have been used to extract probability distribution-dependent features from the predicted states, which then inform a learned ensemble Kalman filter-like update step that assimilates observations~\cite{bach2025learning}. Another class of methods directly learn the optimal (in a statistical sense) filter from data~\cite{levine2022framework,bach2024inverse,mccabe2021learning,bocquet2024accurate,boudier2023data}. By formulating Bayesian inference as a transport problem, yet another class of methods uses generative models for data assimilation. To facilitate the transport necessary for filtering, these methods construct transport maps~\cite{taghvaei2016optimal,taghvaei2021optimal,taghvaei2022optimal} using Knothe–Rosenblatt rearrangements~\cite{spantini2022coupling}, normalizing flows~\cite{chipilski2025exact}, diffusion models~\cite{bao2024score,bao2024ensemble,bao2025nonlinear}, flow matching~\cite{transue2025flow}, and other constructions of optimal transport maps~\cite{al2023nonlinear,al2023optimal}. Although promising, several of these approaches employ neural networks to construct the transport maps. These neural networks require large amounts of data to learn transport maps that ensure good predictive accuracy. 
The issue is further compounded because a new transport map must be learned every time a new measurement is made, by adapting or retraining the neural networks on new data that reflects the evolving dynamics of the system. Therefore, long-trajectory data assimilation with small ensemble sizes using \emph{deep} generative models remains a challenge. Recent versions of these algorithms have begun to address this by performing most of the training offline thereby amortizing the cost over multiple assimilation steps \cite{al2025fast}. 

To address this limitation, the current work explores a different solution. In particular, we consider `\emph{closed-form}' conditional diffusion models for data assimilation. We note that diffusion models have emerged as popular generative models for a wide range of applications~\cite{dhariwal2021diffusion,dasgupta2025conditional,dasgupta2026unifying,lai2025principles}. Diffusion models use a forward process that progressively adds noise to a collection of data points from a target distribution, and learns a score function that is applied iteratively to reverse this process and generate new samples from the target distribution. We are motivated by recent results showing that the score function is analytically tractable~\cite{karras2022elucidating,baptista2025memorization,wang2026error,scarvelis2023closed}. Hence, closed-form (training-free) diffusion models offer unique advantages over neural network-based diffusion models for use in data assimilation. Primarily, the score function can be evaluated exactly, and its evaluation does not require large ensemble sizes.  It is precisely on this premise that we formulate, explore, and carefully evaluate the performance of closed-form conditional diffusion models for data assimilation in this work. 

We note that \cite{bao2024ensemble,transue2025flow} also propose training free diffusion and flow matching models, respectively, for data assimilation with two key differences from the current work. First, the proposed approaches in \cite{bao2024ensemble,transue2025flow} use diffusion and flow matching models to model the prior associated with the probabilistic inverse problem that must be solved for filtering. Second, the proposed approaches in \cite{bao2024ensemble,transue2025flow} account for the measurement through a guidance term that depends on the conditional distribution associated with the measurement conditioned on the state of the system. This requires \textit{a priori} knowledge regarding the distributional form of the measurement process, which may not be available or may be intractable. In contrast, the proposed approach in this work is entirely sample-based and does not require any explicit knowledge about the parametric form of the system. This also makes the current approach suitable for application in black-box settings. 

The remainder of this paper is organized as follows. We set up the data assimilation (filtering) problem in \Cref{sec:background}. \Cref{sec:proposed-approach} introduces the proposed approach. We apply the proposed approach to several data assimilation problems in \Cref{sec:experiments}. Finally, \Cref{sec:conclusion} concludes the paper. 

\section{Background}\label{sec:background}

\subsection{Data assimilation}
Data assimilation is the process of sequentially estimating the states of a stochastic dynamical system as observations become available. At any assimilation step, $k \in \mathbb{Z}^{+}$, the dynamics of the random vector that represents the state, denoted by $\x_{k} \in \mathbb{R}^{d}$, are given by the process model,
\begin{equation}\label{eq:process_model}
    \x_{k} \sim \pi_{\rm proc} (\x_{k} | \x_{k-1}).
\end{equation}
The noisy, and sometimes sparse observations of the state, denoted by $\y_{k} \in \mathbb{R}^D$ are defined in terms of an observation model,
\begin{equation}\label{eq:observation_model}
    \y_{k} \sim \pi_{\rm obs} (\y_{k} | \x_{k}).
\end{equation}
Further, we assume that the initial state is drawn from a known prior distribution 
\begin{equation}\label{eq:initial_state}
    \x_0 \sim \pi(\x_0).
\end{equation}
Together, the process model, the observation model, and the initial state define a probabilistic state-space model (SSM). At this point, we make the following assumptions about the nature of the SSM:
\begin{enumerate}[leftmargin=*] 
    \item \textit{Assumption 1}: The system follows a first-order Markov process. This assumption is used in \Cref{eq:process_model} where the distribution of $\x_{k}$ depends only on the previous state $\x_{k-1}$.
    
    \item \textit{Assumption 2}: The observations are conditionally independent given the state. This assumption is used in \Cref{eq:observation_model} where the distribution of $\y_{k}$ given $\x_{k}$ does not depend on any of the previous measurements.
\end{enumerate} 

\subsection{Bayes Filter}

The goal of Bayesian filtering is to estimate the conditional distribution $\pi(\x_{k} | \hat{\y}_{1:k})$ of the state $\x_k$, given all available observations up to assimilation step $k$, denoted as $\hat{\y}_{1:k}$, where $\hat{\y}_k$ represents a realization of the random variable $\y_k$. Under assumptions 1 and 2, the Bayes filter provides a recursive approach for state estimation. Starting from the distribution of the assimilated state at $k-1$, denoted by $\pi(\x_{k-1} | \hat{\y}_{1:k-1})$, the distribution of the assimilated state at the next step, denoted by $\pi(\x_{k} | \hat{\y}_{1:k})$ can be estimated using the following steps:
\begin{enumerate}
    \item \textbf{Prediction Step:} This step propagates the state estimation forward using the process model \Cref{eq:process_model} and yields the predicted state distribution at assimilation step $k$ via marginalization
    \begin{equation}\label{eq:predict_step}
        \pi(\x_{k} | \hat{\y}_{1:k-1}) = \int \pi_{\rm proc}(\x_{k} | \x_{k-1}) \pi(\x_{k-1} | \hat{\y}_{1:k-1}) \; d\x_{k-1}.
    \end{equation}
    \item \textbf{Update Step:} Upon receiving a new observation $\hat{\y}_{k}$, the observation is incorporated to refine the state estimation by applying Bayes' theorem, where the predicted state distribution (from Eq. \eqref{eq:predict_step}) serves as the prior and the observation model (\Cref{eq:observation_model}) serves as the likelihood,
    \begin{equation}\label{eq:update_step}
        \pi(\x_{k} | \hat{\y}_{1:k}) \propto \pi_{\rm obs}(\hat{\y}_{k} | \x_{k}) \pi(\x_{k} | \hat{\y}_{1:k-1}).
    \end{equation}
\end{enumerate}

This formulation provides an exact framework for state estimation and many popular methods for state estimation, like the Kalman Filter and its variants, and particle filtering methods can be derived from it. However, as we discussed in \Cref{sec:introduction}, these methods are challenged when the process and the observation models are nonlinear, the SSM is non-Gaussian, and when the dimensions of the state and observation vectors are large. 

\section{Proposed approach}\label{sec:proposed-approach}

The proposed methodology is entirely sample-based, which yields the important advantage that no explicit parametric specification of the probability densities in either the process model or the observation model is required. In particular, the approach operates directly on ensembles of samples, thereby avoiding restrictive distributional assumptions and enabling applicability to nonlinear and non-Gaussian systems.

\subsection{Overall approach}

We assume that at assimilation step $k-1$ we have access to $N$ samples representing the filtering distribution, 
\ie samples drawn from $\pi(\x_{k-1} | \hat{\y}_{1:k-1})$. The objective is to construct an algorithm that, given these $N$ samples, produces $N$ samples from the filtering distribution 
at the current assimilation step $k$, namely samples from $\pi(\x_k | \hat{\y}_{1:k})$. This algorithm is derived in the following development and is described in detail in  \Cref{alg:proposed-diffusion}. 

The prediction step in the proposed framework coincides with the prediction step used in standard sample-based data assimilation methods. Specifically, each of the $N$ samples from the assimilated state at time $k-1$ is propagated through the process model. This yields $N$ predicted samples distributed according to the forecast density $
\pi(\x_k | \hat{\y}_{1:k-1})$, which represents the prior distribution at time $k$ before incorporating the new measurement; see \Cref{eq:update_step}.

For the update step, we employ a conditional diffusion model. To simplify notation, we introduce the following shorthand: the random vector $\x_k \mid \hat{\y}_{1:k-1}$ is denoted by $\x$; the random measurement vector $\y_k$ is denoted by $\y$; the realized measurement $\hat{\y}_k$ is denoted by $\hat{\y}$; and the posterior random vector $\x_k \mid \hat{\y}_{1:k}$ is denoted by $\x \mid \hat{\y}$. Under this notation, the Bayesian update step \Cref{eq:update_step} can be rewritten as
\begin{equation}\label{eq:update_step1}
    \pi(\x \mid \hat{\y}) \propto \pi_{\rm obs}(\hat{\y} \mid \x) \, \pi(\x).
\end{equation}
Accordingly, the update problem can be reformulated as follows: given $N$ samples $\x^{(i)}$, $i=1,\ldots,N$, drawn from the prior distribution $\pi(\x)$, together with the capability to generate samples from the observation model and the realized measurement $\hat{\y}$, construct $N$ samples from the posterior distribution $\pi(\x \mid \hat{\y})$ defined in \Cref{eq:update_step1}.

To achieve this, we first apply the observation model to each prior sample $\x^{(i)}$ to generate the corresponding synthetic observations $\y^{(i)}$, thereby forming paired samples $(\x^{(i)}, \y^{(i)})$, $i=1,\ldots,N$. These paired samples are then used in a conditional diffusion model to generate $N$ samples from the posterior distribution $\pi(\x \mid \hat{\y})$, thereby completing the update step.

\subsection{Closed-form conditional diffusion model}

The principal idea behind a diffusion model is a forward process that progressively adds noise to a collection of data points, which are sampled from an underlying data distribution, and a reverse process that starting from noise progressively removes noise to yield samples from the data distribution~\cite{lai2025principles}. A key component of the reverse process is the score function~\cite{song2019generative}, which is usually parameterized using neural networks. However, the score function is analytically tractable in some cases~\cite{karras2022elucidating,baptista2025memorization,dasgupta2026unifying}, which is utilized to formulate closed-form diffusion models~\cite{scarvelis2023closed}. Below we define the forward and reverse processes associated with the conditional diffusion model, and then extract the score function using the empirical joint distribution of the paired samples $(\x^{(i)}, \y^{(i)})$, $i=1,\ldots,N$. 

\subsubsection{Forward process}
We first define a diffusion process that maps samples from the desired $\pi(\bm{x} | \y)$ to samples from a multivariate Gaussian distribution with zero mean and large variance. To accomplish this, we define a pseudo-time coordinate $t \in (0,1)$, and introduce the pseudo-time dependent density $\pi(\x,t | \y)$ that satisfies the diffusion equation~\cite{dasgupta2026unifying}, 
\begin{equation}
    \frac{\partial \pi(\x,t | \y)}{\partial t}
    = \frac{\gamma(t)}{2} \Delta \pi(\x, t | \y)
\end{equation}
along with the initial condition $\pi(\x,0 | \y) = \pi(\x | \y)$. 
The solution to this equation is given by 
\begin{equation}\label{eq:pdesol}
    \pi(\x, t | \y)
    = \int_{\mathbb{R}^d} g_{\sigma(t)}(\x - \x') \pi(\x'|\y) \mathrm{d}\x' 
\end{equation}
where the Gaussian kernel,
\begin{equation}\label{eq:gaussian_kernel}
     g_{\sigma}(\x)
    \equiv \mathcal{N}(\x; \bm{0}, \sigma^2\mathbb{I}_d)
    = \frac{1}{(2\pi \sigma^2)^{d/2}} \exp \left(-\frac{\lVert \x \rVert^2_2}{2 \sigma^2}\right)
\end{equation}
and $\sigma(t)$ is given by 
\begin{equation}
    \sigma^2(t) = \int^t_0 \gamma(s) \mathrm{d}s.
\end{equation}
The schedule for the parameter $\gamma(t)$ is selected such that $\sigma(0)=0$ and $\sigma(1) \gg 1$. In practice, we prescribe a schedule for $\sigma (t)$ where we linearly interpolate it between $\sigma_{\rm max}$ and $0$, \ie we choose $\sigma(t) = t\sigma_{\rm max}$. 

\subsubsection{Reverse process}
Next, we define the backward pseudo-time $\tau \equiv 1-t$, and define a random vector $\x(\tau)$. It can be shown that if samples $\x^{(i)}(0) \sim \mathcal{N}(\bm{0}, \sigma^2 (1) \mathbb{I}_n)$ and are evolved via 
\begin{equation}\label{eq:reverse}
    \frac{\mathrm{d} \x^{(i)}(\tau)}{\mathrm{d} \tau} = \frac{\gamma(t)}{2} \bm{s}(\x^{(i)}(\tau), t|\hat{\y}), 
\end{equation}
where $t = 1-\tau$, and the score function is defined as,
\begin{equation}
    \bm{s}(\x, t|\y) = \nabla \log \pi(\x, t|\y),
\end{equation}
then $\x^{(i)}(1) \sim \pi(\x|\hat{\y})$~\cite{dasgupta2026unifying}.  The results above describe how we can transport samples drawn from a simple Gaussian distribution to samples from the desired conditional distribution. However, this requires explicit knowledge of the score function.  Next, we show how the score function can be determined using the samples $(\x^{(i)}, \y^{(i)}) \sim \pi(\x, \y)$. 

\subsubsection{Closed-form for the score function}
We use the paired samples $(\x^{(i)}, \y^{(i)})$, $i=1,\ldots,N$, to write the empirical approximation to the joint probability density function (pdf),
\begin{equation}
    \pi(\x, \y)
    \approx \frac{1}{N} \sum^{N}_{i=1} \delta(\x - \x^{(i)}) \delta(\y - \y^{(i)}),
\end{equation}
where $\delta(\cdot)$ denotes the Dirac delta function. We then use kernel density estimation (KDE) to obtain 
a smooth approximation,
\begin{equation}\label{eq:kde_joint}
    \pi(\x, \y)
    \approx \frac{1}{N} \sum^{N}_{i=1} g_{\sigma_x}(\x - \x^{(i)}) g_{\sigma_y}(\y - \y^{(i)}), 
\end{equation}
where $\sigma_x$ and $\sigma_y$ represent the kernel widths along the state and the measurement coordinates, respectively. Given this approximation for the joint density, we now evaluate an explicit form for the pseudo-time dependent conditional probability density function. From \Cref{eq:pdesol} we have, 
\begin{align}\label{eq:cond_density_exp}
    \pi(\x, t | \y)
    =& \int_{\mathbb{R}^d} g_{\sigma(t)}(\x - \x') \pi(\x'|\y) d\x' \nonumber\\
    =& \int_{\mathbb{R}^d} g_{\sigma(t)}(\x - \x') \frac{\pi(\x', \y)}{\pi(\y)} d\x' \nonumber\\
    =& \frac{1}{\pi(\y)} \int_{\mathbb{R}^d} g_{\sigma(t)}(\x - \x') \left[ \frac{1}{N} \sum^{N}_{i=1} g_{\sigma_x}(\x' - \x^{(i)}) g_{\sigma_y}(\y - \y^{(i)})   \right] d\x' \nonumber\\
    =& \frac{1}{N\pi(\y)} \sum^{N}_{i=1} g_{\sigma_y}(\y -  \y^{(i)}) \int_{\mathbb{R}^d} g_{\sigma(t)}(\x - \x') g_{\sigma_x}(\x' - \x^{(i)}) d\x' \nonumber\\
    =& \frac{1}{N\pi(\y)} \sum^{N}_{i=1} g_{\sigma_y}(\y- \y^{(i)}) g_{\bar{\sigma}(t)}(\x - \x^{(i)})
\end{align}
where $\bar{\sigma}(t) = \sqrt{\sigma^2(t)+\sigma_x^2}$. In the development above, in the second line we have used the definition of a conditional density, in the third line we have employed \Cref{eq:kde_joint}, and in the fifth line we have used a result that applies to the convolution of two Gaussian kernels and is proven in \Cref{appsubsec:convolution-Gaussians}. 

Using the approximation for the conditional density from \Cref{eq:cond_density_exp}, we can evaluate the score function as follows,
\begin{align}\label{eq:score_expression}
    \bm{s}(\x, t | \y) 
    =& \nabla_{\x} \log \pi(\x, t | \y) \nonumber\\
    =& \frac{\nabla_{\x} \pi(\x, t | \y)}{\pi(\x, t | \y)} \nonumber\\
    =& \frac{\sum^{N}_{i=1} g_{\sigma_y}(\y - \y^{(i)})\nabla_{\x} g_{\bar{\sigma}(t)}(\x -  \x^{(i)})}{\sum^{N}_{j=1} g_{\sigma_y}(\y - \y^{(j)}) g_{\bar{\sigma}(t)}(\x -  \x^{(j)})}\nonumber\\
    =& \frac{\sum^{N}_{i=1} g_{\sigma_y}(\y - \y^{(i)}) g_{\bar{\sigma}(t)}(\x -  \x^{(i)}) \left(\frac{\x^{(i)} - \x}{\bar{\sigma}^2(t)}\right) }{\sum^{N}_{j=1} g_{\sigma_y}(\y - \y^{(j)}) g_{\bar{\sigma}(t)}(\x -  \x^{(j)})}\nonumber \\
    =& \sum^{N}_{i=1}  \bar{w}^{(i)}(\x, \y, t) \frac{\x^{(i)} - \x}{\bar{\sigma}^2(t)}, 
\end{align}
where the weights 
\begin{equation}\label{eq:def_weights}
\bar{w}^{(i)}(\x, \y, t) = \frac{g_{\sigma_y}(\y - \y^{(i)}) g_{\bar{\sigma}(t)}(\x -  \x^{(i)})}{\sum_{j = 1}^N g_{\sigma_y}(\y - \y^{(j)}) g_{\bar{\sigma}(t)}(\x -\x^{(j)})}.    
\end{equation}
In deriving the expression for the score function (\Cref{eq:score_expression}), in the first line we have used the definition of the score function, in the second line we have used the definition of a conditional density, in the third line we have employed the expression for the conditional density from \Cref{eq:cond_density_exp} and recognized that the gradient is with respect to the $\x$ coordinates only, in the fourth line we have made use of a property of the Gaussian kernel that is derived in \Cref{appsubsec:gradient_Gaussian}, and in the fifth line we have used the definition of the weights \Cref{eq:def_weights}.

\subsection{Putting it all together}\label{sec:all_together}
\Cref{alg:proposed-diffusion} summarizes the final algorithm for a single step of the data assimilation process. We note that when implementing this algorithm we subtract the mean from the paired data, $(\x^{(i)}, \y^{(i)}), i=1,\ldots,N$, to ensure that its zero-mean and then normalize it between [-1,1] before sampling commences (step 3 of \Cref{alg:proposed-diffusion}). The observation $\hat{\y}_k$ is also appropriately transformed. Significantly, the kernel bandwidths $\sigma_x$ and $\sigma_y$ in \Cref{eq:kde_joint} are chosen relative to this normalized scale. Step 5 in \Cref{alg:proposed-diffusion}, which involves integrating \Cref{eq:reverse}, can be accomplished using any suitable numerical integration procedure and is easily parallelizable across the ensemble.  We use an adaptive, explicit Runge-Kutta method of order 5(4), available through \texttt{SciPy}'s~\cite{2020SciPy-NMeth} \texttt{solve\textunderscore ivp} routine to integrate \Cref{eq:reverse} in parallel for the entire ensemble. Finally, the samples are re-normalized after sampling. 
\begin{algorithm}[!t]
\caption{Proposed algorithm for a single data assimilation step}\label{alg:proposed-diffusion}
\KwIn{Samples $\x^{(i)}_{k-1} \sim \pi(\x_{k-1} \mid \hat{\y}_{1:k-1}), i = 1, \ldots, N$, measurement $\hat{\y}_{k}$, and kernel bandwidths $\sigma_x$ and $\sigma_y$}
\KwOut{Samples $\x^{(i)}_{k} \sim \pi(\x_{k} \mid \hat{\y}_{1:k}), i = 1, \ldots, N$}
Generate $ \x^{(i)} \sim \pi_{\rm proc}(\cdot \mid \x_{k-1}^{(i)}), i = 1, \ldots, N$ using the process model\\ \hphantom{Generate $ \x^{(i)} \sim \pi_{\rm proc}(\cdot \mid \x_{k-1}^{(i)}), i = 1, \ldots, N$ using the abc} \Comment{Prediction Step}\;
Generate synthetic measurements $ \y^{(i)} \sim \pi_{\rm obs} (\cdot \mid \x^{(i)}), i = 1, \cdots, N$ using the measurement model\;
Select $ \x^{(i)}(0) \sim \mathcal{N}(\bm{0}, \sigma^2(1) \mathbb{I}_n), i = 1,\cdots, N$ \Comment{Initial conditions for \Cref{eq:reverse}}\;
\For{$i=1,\dots,N$}{
         Integrate \Cref{eq:reverse} in the interval $(0, 1)$ to obtain $\x^{(i)}(1)$ \Comment{Sampling}
}
\Return{$\x^{(i)}_{k} \gets \x^{(i)}(1)$} \Comment{Update Step}
\end{algorithm}

\section{Numerical Experiments}\label{sec:experiments}
In this section, we evaluate the performance of the diffusion model–based data assimilation method against standard ensemble-based filtering approaches. We begin by describing the common experimental setup used across all simulations, including model propagation, observation generation, initialization procedures, hyperparameter selection, and evaluation metrics. We then present results for three benchmark systems: Lorenz–63~\cite{lorenz1963deterministic}, Lorenz–96 in 10 dimensions, and Lorenz–96 in 20 dimensions~\cite{lorenz1996predictability}. All of these problems exhibit nonlinear and chaotic dynamics, and the associated observation operators are also nonlinear. In addition, the Lorenz–63 problem is constructed such that the true density of the estimated state is bimodal for a significant portion of the simulation~\cite{al2025fast}, thereby posing a challenge to most data assimilation strategies. The Lorenz–96 configurations are designed so that the dimension of the inferred state is reasonably large, further increasing the level of difficulty.

\subsection{Experimental Set Up}

We describe how the state-space model defined in \Cref{sec:background} is instantiated for the numerical experiments. 

\paragraph{Process model}
In the examples considered in this study the underlying state of the system, denoted by $\bm{u}(t) \in \mathbb{R}^d$ is continuous in physical time $t \in (0,T)$, and is required to satisfy the set of first-order ODEs given by 
\begin{equation}\label{eq:exp_dynamics}
    \frac{\mathrm{d}\bm{u}}{\mathrm{d}t} =  \bm{f}(\bm{u},t).
\end{equation}
Observations are made at time instances $t_k = k \times \Delta t$ separated by the observation interval $\Delta t$. The state at these time instances, which is denoted by $\x_k \equiv \bm{u}(t_k)$ is assimilated with the help of the observations. The explicit form of the process model is 
\begin{equation}
    \x_k = \bm{\Psi}(\x_{k-1}) + \bm{\epsilon}_k,
\end{equation}
where the operator $\bm{\Psi}$ maps the state at time $t_{k-1}$ to the state at time $t_k$ by using the former as an initial condition and integrating the ODE \Cref{eq:exp_dynamics}. Further, $\bm{\epsilon}_k \sim \mathcal{N}(\bm{0}, \sigma_\epsilon^2 \mathbb{I}_d)$ is the additive process noise.  

\paragraph{Observation model} Observations are generated according to an observation operator
\begin{equation}\label{eq:exp_obs}
    \y_k = \bm{h}(\x_k, \bm{\eta}_k),
\end{equation}
where $\bm{\eta}_k$ denotes the observation noise.

\paragraph{Benchmark Filters}
The proposed diffusion model-based filter is compared against the following ensemble-based data assimilation methods:
\begin{enumerate}[leftmargin=*,itemsep=0pt]
    \item A stochastic Ensemble Kalman Filter (EnKF)~\cite{katzfuss2016understanding} based on a Gaussian approximation to the filtering distribution with empirical updates to the mean and covariance. 
    \item Sequential Importance Resampling (SIR)~\cite{doucet2000sequential} particle filter with multinomial resampling.
\end{enumerate}

\paragraph{Simulation Procedure} Given a specified process and observation model, each numerical experiment is conducted according to the following procedure:
\begin{enumerate}[leftmargin=*,itemsep=0pt]
    \item Generate reference states: The initial true state is sampled as $\x^*_0 \sim \mathcal{N}(\bm{0}, \mathbb{I}_d)$. The reference states $\{\x^*_k\}_{k=0}^K$ are then obtained by integrating \Cref{eq:exp_dynamics} forward in time without process noise.

    \item Generate realized observations: At each assimilation step $k$, observations are generated by applying \Cref{eq:exp_obs} to the reference states $\{\x^*_k\}_{k=1}^K$. This yields the set of observations $\{\hat{\y_k}\}_{k=1}^K$.

    \item Initialize the ensemble: We assume that we know the distribution of the true state at $k=0$. Thus, in the data assimilation process, the initial ensemble is constructed as $\x_0^{(i)} \sim \mathcal{N}(\x^*_0, \mathbb{I}_d),
    \: i = 1,\dots,N$, where $N$ is the ensemble size.    

    \item Run filtering: Each filter is applied for $K$ assimilation steps using the realized observations $\{\hat{\y_k}\}_{k=1}^K$ (from step 2).

    \item Evaluate Performance: A system-dependent performance metric is computed at each assimilation step and averaged over all the assimilation steps. This includes the Wasserstein distance between a benchmark and the assimilated state for a given method for the Lorenz-63 system, and the root mean square error (RMSE) between the the true state and the mean of the assimilated state for a given method for the Lorenz-96 problem. 
\end{enumerate}

\paragraph{Selection of hyperparameters}  We select $\sigma_{\rm \max} = 5$ for all experiments and observe that the performance of the proposed approach is fairly robust to this choice. Therefore, the only tunable parameters in the proposed method are the bandwidths $\sigma_x$ and $\sigma_y$ in \Cref{eq:kde_joint}. The optimal values for these parameters are determined by performing a grid search and are reported for each case. 
For each experimental configuration, we perform \(S=10\) independent simulations with independently generated true states and observations, and report performance metrics averaged over these simulations. 
We also evaluate the performance of all filters at ensemble sizes given by $N\in \{20, 50, 100, 250, 500, 1000\}$.

\subsection{Lorenz-63}
The Lorenz-63 model is a widely used benchmark in data assimilation due to its nonlinear, low-dimensional chaotic dynamics, and we adapt this example from \cite{al2025fast}. The dynamics are governed by \Cref{eq:exp_dynamics}, where the components of the vector on the right hand side are given by
\begin{equation}
    f_1 = \sigma (u_2-u_1), \quad
    f_2 = \rho u_1 - u_2 - u_1 u_3, \quad 
    f_3 = u_1 u_2 - \beta u_3. 
\end{equation}
The values of the parameters in these equations and the numerical scheme used to integrate these equations are reported in \Cref{tab:L63-settings}. The observation operator is defined by 
\begin{equation}
    y_1 = x_3 + \eta, 
\end{equation}  
where the additive noise and the observation time step ($\Delta t$) are reported in \Cref{tab:L63-settings}.

\begin{table}[h]
    \centering
    \caption{Lorenz-63 Experiment Settings}
    \label{tab:L63-settings}
    \begin{tabular}{ll}
    \toprule
    \textbf{Category} & \textbf{Values} \\
    \midrule
    System Parameters & $\sigma = 10, \; \rho = 28, \; \beta = 8/3$ \\
    Assimilation Schedule & $K=100, \; \Delta t=0.1$ \\
    Process Integration Scheme & Forward Euler ($dt=0.01$) \\
    Process Noise & $\bm{\epsilon} \sim \mathcal{N}(\bm{0}, 0.01^2\mathbb{I}_3)$\\
    Observation Noise & $\eta \sim \mathcal{N}(0, 0.5^2)$ \\
    \bottomrule
    \end{tabular}
\end{table}

Due to partial observations and nonlinear dynamics, the assimilated distribution exhibits significant non-Gaussian features, including many instances when it is bimodal. Consequently, computing the filtering error relative to the true reference states $\bm{x}^*_k$ is insufficient to assess distribution accuracy for any method. Instead, we evaluate the discrepancy between the estimated assimilated distribution for a method and a high-fidelity reference distribution. For computing the high-fidelity distribution we utilize an SIR filter with $N_{\rm true}=100,000$ particles. It is well know that as the ensemble size of the SIR filter increases, it converges towards the true distribution, and is therefore a good reference solution for low-dimensional systems. 

\begin{table}[!b]
\centering
\caption{For each ensemble size $N$, the optimal kernel bandwidth parameters $(\sigma_x, \sigma_y)$ selected for the conditional diffusion-based filter, the range of sampling steps used by the adaptive numerical integration scheme (see \Cref{sec:all_together}), and the corresponding time-averaged Wasserstein error $\mathcal{E}_{W_2}$ are reported for the Lorenz–63 system. The time-averaged Wasserstein errors for the EnKF and SIR filters are also reported. All errors are also averaged over $S=10$ independent simulations. We indicate the best performing filter using bold fonts.}
\label{tab:L63-Results}
\begin{tabular}{ccccccc}
\toprule
\multirow{2}{*}{\makecell[c]{\textbf{Ensemble size}\\N}} & \multicolumn{2}{c}{\makecell{\textbf{Optimal Bandwidth}}} &\multirow{2}{*}{\makecell[c]{\textbf{Range of}\\\textbf{sampling steps}}} &
\multicolumn{3}{c}{\textbf{Average $\mathcal{E}_{W_2}$}} \\
\cmidrule(rl){2-3}
\cmidrule(rl){5-7}
& {$\sigma_x$} & {$\sigma_y$} & & {Diffusion} & {EnKF} & {SIR} \\
\midrule
~~20   & 0.200 & 0.50 & ~9 -- 12 & \textbf{12.809} & 14.768 & 17.440 \\
~~50   & 0.100 & 0.50 & 12 -- 16 & ~\textbf{9.774}  & 13.773 & 17.867 \\
~100  & 0.100 & 0.25 & 10 -- 13 & ~\textbf{8.474}  & 13.768 & 17.400 \\
~250  & 0.050 & 0.25 & 11 -- 15 & ~\textbf{6.553}  & 12.017 & 16.882 \\
~500  & 0.025 & 0.25 & 12 -- 16 & ~\textbf{6.233}  & 11.865 & 15.332 \\
1000 & 0.025 & 0.25 & 13 -- 17 & ~\textbf{5.744}  & 12.944 & 14.851 \\
\bottomrule
\end{tabular}
\end{table}
We let $\{\x^{(i)}_k\}_{i=1}^N$ denote the ensemble produced by a given filter at assimilation step $k$, and let $\{\tilde{\x}^{(i)}_k\}_{i=1}^{N_{\rm true}}$ denote the reference SIR ensemble. Their corresponding empirical approximations of the probability density distributions are as follows,
\begin{equation}
    \mu_k = \frac{1}{N} \sum^{N}_{i=1} \delta (\x-\x^{(i)}_k), 
    \quad \tilde{\mu}_k = \frac{1}{N_{\rm true}} \sum^{N_{\rm true}}_{i=1} \delta (\x-\tilde{\x}^{(i)}_k). 
\end{equation}
We quantify the discrepancy between these distributions using the Wasserstein-2 distance and average it over all assimilation steps. This error is given by,
\begin{equation}
    \mathcal{E}_{W_2} = \frac{1}{K} \sum_{k=1}^K W_2(\mu_k, \tilde{\mu}_k).
\end{equation}
For the three methods considered in this study, this error is reported in \Cref{tab:L63-Results}. From this table, we conclude that for all ensemble sizes, the proposed approach outperforms the EnKF and SIR methods. Further, the error in this method reduces with increasing ensemble size, whereas for the EnKF and SIR methods, the error does not appear to reduce with increasing ensemble size (for the range reported in this table). Further, in \Cref{tab:L63-Results}, we also report the range of the number of adaptive steps taken when integrating \Cref{eq:reverse}. \Cref{tab:L63-Results} shows that integrating \Cref{eq:reverse} typically requires only a few steps, and thus only a small number of evaluations of the right hand side of \Cref{eq:reverse}. 


\begin{figure}[!t]
    \centering
    \begin{subfigure}{0.48\textwidth}
        \centering
        \includegraphics[width=\linewidth]{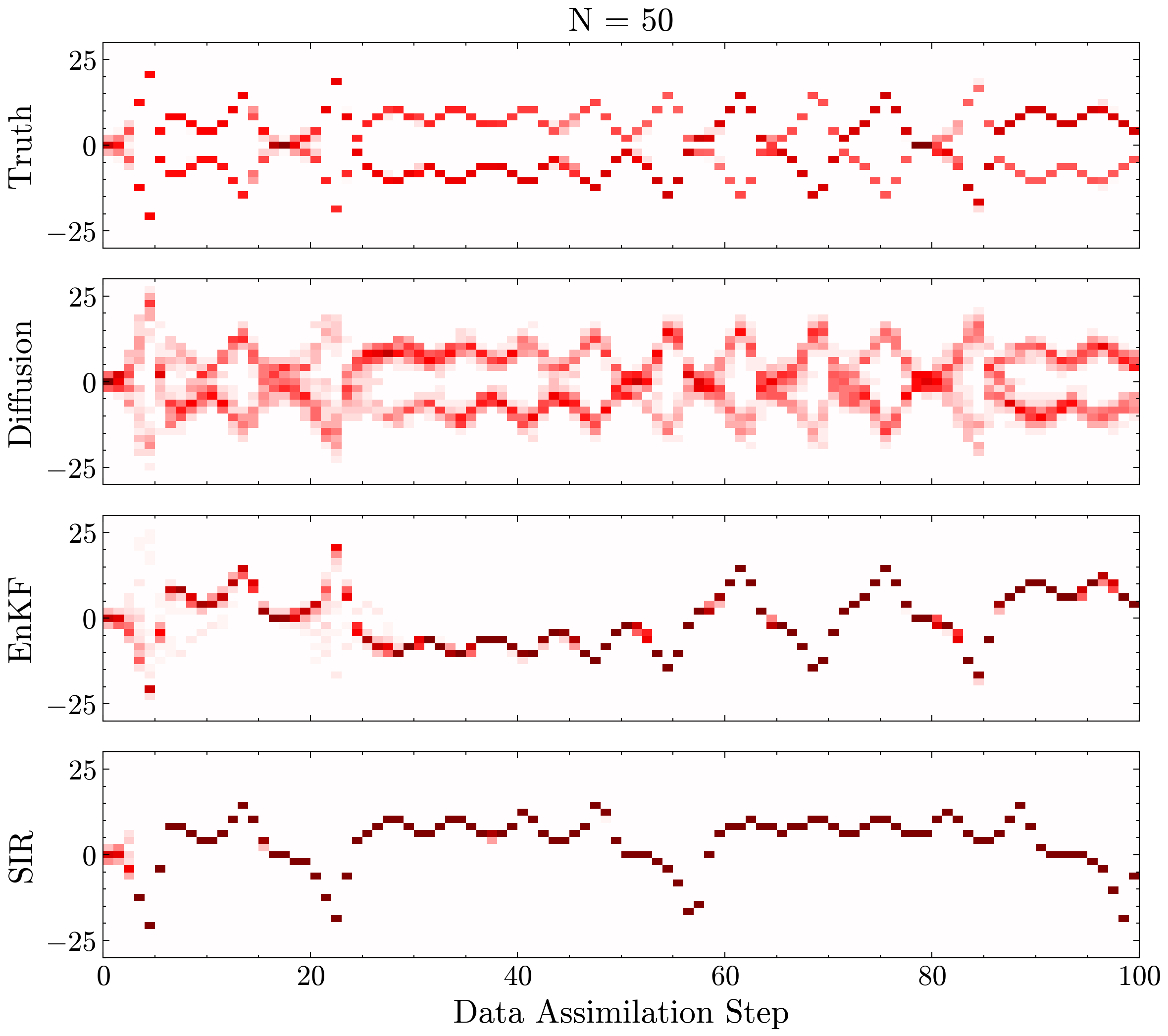}
        \caption{$N=50$ Simulation=3}
    \end{subfigure}
    \hfill
    \begin{subfigure}{0.48\textwidth}
        \centering
        \includegraphics[width=\linewidth]{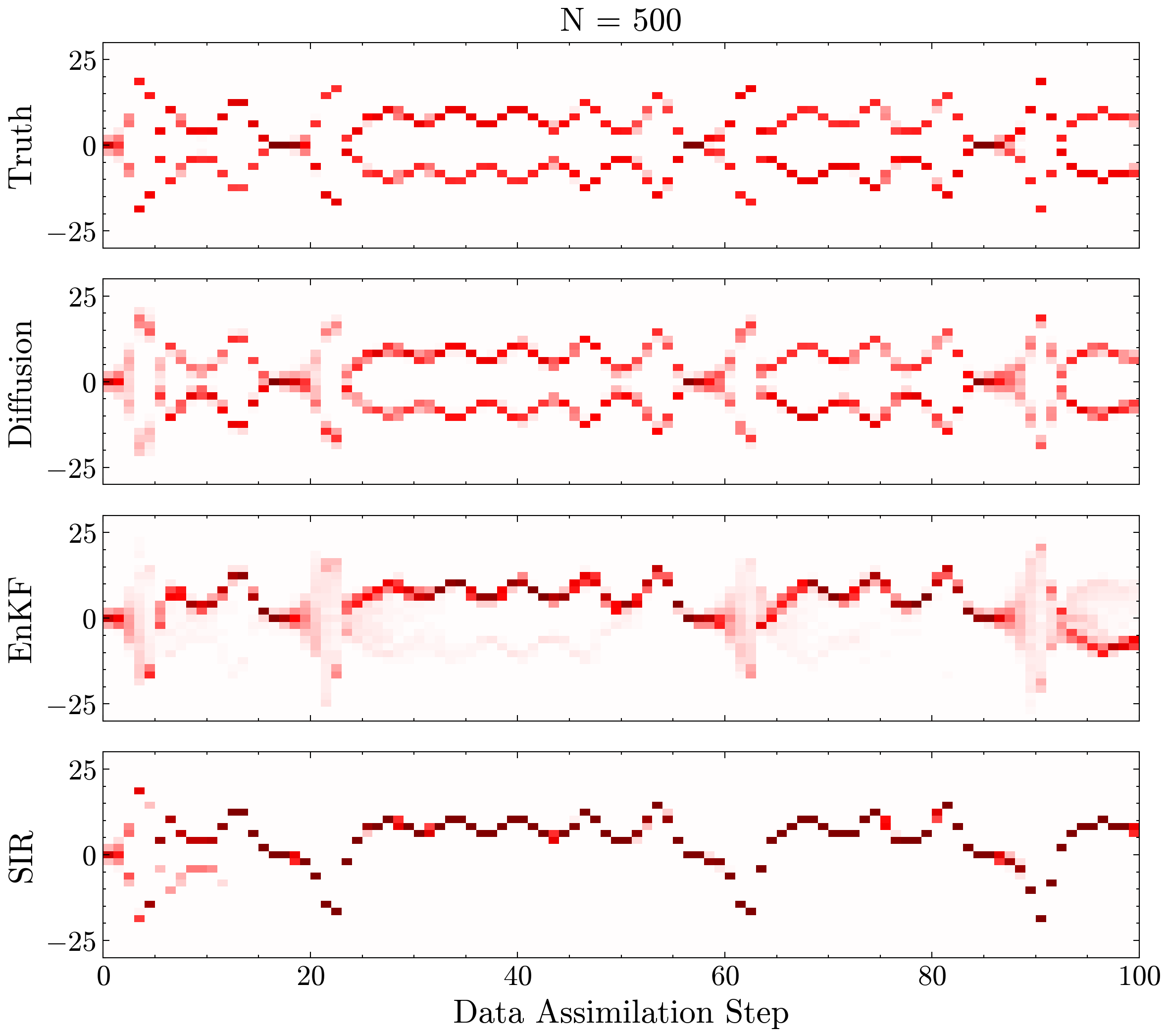}
        \caption{$N=500$, Simulation=6}
    \end{subfigure}
    \caption{Evolution of the filtering distribution of $x_1$ (vertical axis) for the Lorenz-63 system over $K=100$ assimilation steps (horizontal axis). Each column corresponds to a different ensemble size and simulation. Within each column, the rows show (top to bottom): the reference distribution obtained using SIR with $N_{\rm true}=100{,}000$ particles (Truth), the filtering distribution obtained with ensemble size $N$ using three different filters. The color intensity corresponds to the estimated probability density, with darker regions indicating higher density.}
    \label{fig:L63-densities}
\end{figure}

\begin{figure}[t]
    \centering
    \includegraphics[width=1.0\textwidth]{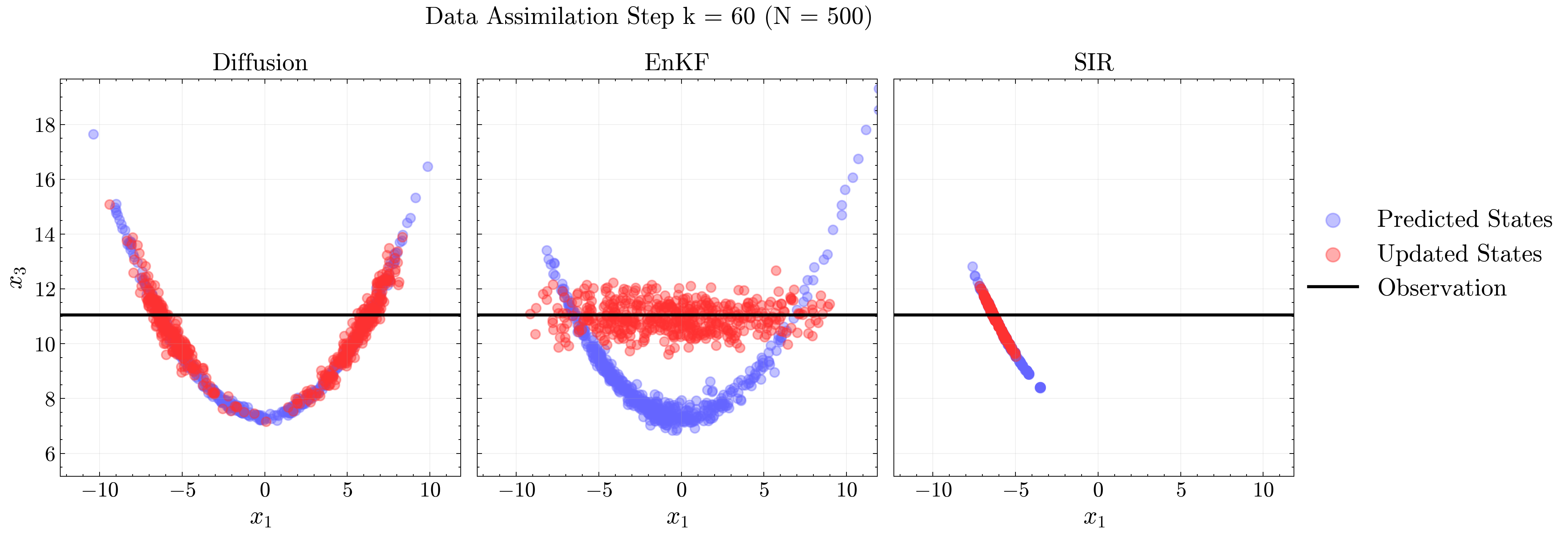}
    \caption{Predicted and updated states at assimilation step $k=60$ and for $N=500$ with three different filters. Red particles represent the states after the prediction step, blue particles represent the states after the update step, and the black line indicates the observation.}
    \label{fig:L63-step}
\end{figure}

The qualitative aspects of the performance of the filtering models can be gleaned from \Cref{fig:L63-densities}, where we have plotted the assimilated distribution for $x_1$ as a function of assimilation step obtained from the benchmark and the three models for two ensemble sizes. Notably, the conditional diffusion model-based filter preserves the bimodal posterior structure, even with small ensemble sizes like $N=50$. In this regime, the EnKF and SIR filters fail for different reasons. The EnKF enforces a Gaussian posterior approximation through its linear updates, which inherently limits its ability to represent multimodal distributions. As a result, the posterior is effectively smoothed into a unimodal form, even when the true filtering distribution is bimodal. On the other hand, the SIR filter exhibits weight degeneracy for small ensemble sizes, due to which the filtering distribution concentrates on a single mode and fails to capture the full posterior. When considering a larger ensemble size, we observe that the distribution for the diffusion model becomes less spread out while maintaining the bimodal structure, whereas for the EnKF and the SIR models we do not observe any qualitative improvement in the distribution. 

These behaviors of the three models are clearly visible in \Cref{fig:L63-step}, which shows the results for the \supth{60} assimilation step. In this plot, which is confined to the $x_1$-$x_3$ in the state space, the blue particles denote the samples obtained at the end of the prediction step, which serve to define the prior distribution for each method. The red particles denote the updated samples which define the assimilated state at the end of this step. For the diffusion filter, the assimilated samples appear to concentrate along a distribution with two modes, whereas for the EnKF filter, the samples are smeared around these two modes, and for the SIR filter only one of the two modes is captured. Overall, we see that the proposed conditional diffusion approach more accurately captures the bimodal structure of the true filtering distribution for this Lorenz-63 example.

\subsection{Lorenz-96}
The Lorenz-96 model is a set of ordinary differential equations that is often used to represent atmospheric dynamics. The dynamics are governed by \Cref{eq:exp_dynamics}, where the components of the vector on the right hand side are given by
\begin{equation}
    f_i = (u_{i+1} - u_{i-2}) u_{i-1} - u_i + F , \quad i=1,...,d
\end{equation}
The components are cyclic i.e. $u_{-1} = u_{d-1}$, $u_0 = u_{d}$, and $u_1 = u_{d+1}$. 

For this example, we consider the following nonlinear observation operator, which is adapted from \cite{bao2024ensemble},
\begin{equation}
    \y = \arctan(\x) + \bm{\eta}. 
\end{equation}
\Cref{tab:L96-settings} shows the specifics of the Lorenz-96 system considered in this example. As part of this example, we consider two different dimensions of the state vector with $d=10$ and $20$.

\begin{table}[t]
    \centering
    \caption{Lorenz-96 Experiment Settings}
    \label{tab:L96-settings}
    \begin{tabular}{ll}
    \toprule
    \textbf{Category} & \textbf{Values} \\
    \midrule
    System Parameters & $F=8$ \\
    Assimilation Schedule & $K = 500, \; \Delta t = 0.1$ \\
    Process Integration scheme & RK4 ($dt=0.01$) \\
    Process Noise & $\bm{\epsilon} \sim \mathcal{N}(\bm{0}, 0.01^2\mathbb{I}_d)$\\
    Observation Noise & $\bm{\eta} \sim \mathcal{N}(\bm{0}, 0.5 \mathbb{I}_d)$ \\
    \bottomrule
    \end{tabular}
\end{table}

Unlike the previous example of the Lorenz-63 system, the Lorenz-96 system does not exhibit multimodal behavior with our choice of the observation operator. Therefore, in order to evaluate the performance of the filter methods, we calculate the RMSE from the true state and average across all assimilation steps. This error is defined as,
\begin{equation}
    \mathcal{E}_{\rm RMSE} = \frac{1}{K}\sum_{k=1}^{K}\frac{\lVert \x^*_k - \bar{\x}_k \lVert_2}{\sqrt{d}}
\end{equation}
where $\bar{\x}_k = \frac{1}{N} \sum^N_{i=1} \x^{(i)}_k$ is the component-wise mean of the estimated state. 

\subsubsection{Lorenz-96 in 10-dimensions}

\begin{figure}[!t]
    \centering
    \includegraphics[width=1.0\textwidth]{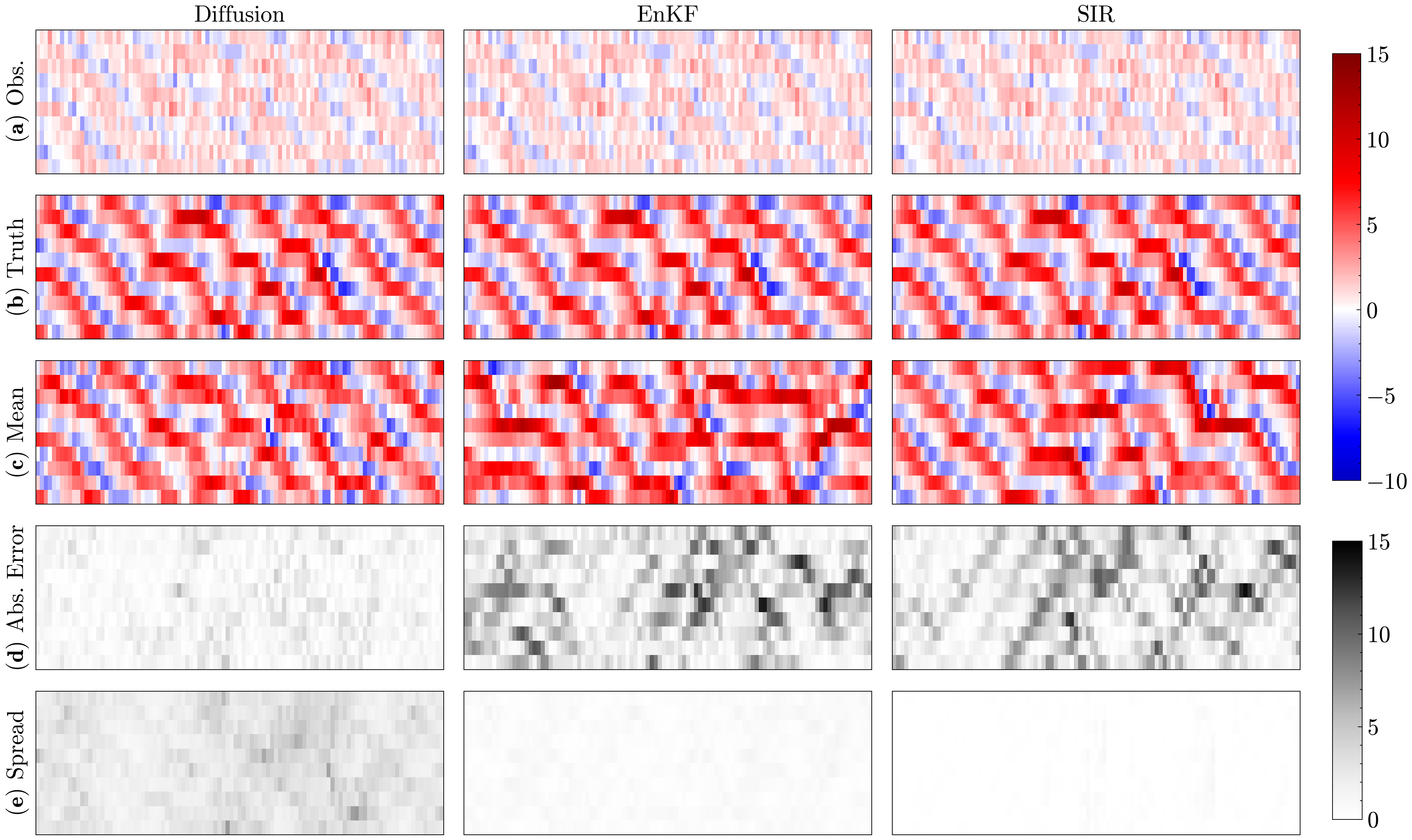}
    \caption{Visualization the performance of different filters with ensemble size $N=100$ for one test trajectory of the 10-dimensional Lorenz-96 system over data assimilation steps 300-400. In each plot, the vertical axis corresponds to the states, and the horizontal axis corresponds to the data assimilation step. Rows 1 through 5 correspond to the observations, true states,  estimated mean of the assimilated states, absolute error of mean with respect to the ground truth, and estimated ensemble spread (one standard deviation about the estimated mean), respectively. Columns 1 through 3 correspond to the conditional diffusion model-based, EnKF, and SIR filters, respectively.}
    \label{fig:L96-10D}
\end{figure}

\begin{table}[!b]
\centering
\caption{Optimal kernel bandwidths $(\sigma_x, \sigma_y)$, the range of sampling steps used by the adaptive numerical integration scheme, and the corresponding time-averaged RMSE error $\mathcal{E}_{RMSE}$ of the conditional diffusion model-based filter with varying ensemble sizes $N$ for the 10-dimensional Lorenz–96 system. The time-averaged RMSE errors for the EnKF and SIR filters are also reported. All errors are also averaged over $S=10$ independent simulations. We indicate the best performing filter using bold fonts.}
\label{tab:L96-10D-Results}
\begin{tabular}{ccccccc}
\toprule
\multirow{2}{*}{\makecell[c]{\textbf{Ensemble size}\\N}} & \multicolumn{2}{c}{\textbf{Optimal Bandwidth}} &  \multirow{2}{*}{\makecell[c]{\textbf{Range of}\\\textbf{sampling steps}}}&\multicolumn{3}{c}{\textbf{Average $\mathcal{E}_{RMSE}$}} \\
\cmidrule(rl){2-3}
\cmidrule(rl){5-7}
& {$\sigma_x$} & {$\sigma_y$} & & {Diffusion} & {EnKF} & {SIR} \\
\midrule
20   & 0.20  & 1.00  & 13 --17 & \textbf{3.073} & 4.904 & 4.806 \\
50   & 0.20  & 0.75 & 14 --17 & \textbf{2.049} & 4.664 & 4.803 \\
100  & 0.20  & 0.50 & 16 --19 & \textbf{1.688} & 4.023 & 4.675 \\
250  & 0.10   & 0.50  & 16 --19 & \textbf{1.144} & 2.195 & 4.742 \\
500  & 0.10   & 0.50  & 17 --19 & 1.076 & \textbf{0.729} & 4.732 \\
1000 & 0.05  & 0.50  & 17 --19 & 0.952 & \textbf{0.592} & 4.563 \\
\bottomrule
\end{tabular}
\end{table}
First, we consider the Lorenz-96 systems in 10 dimensions. \Cref{tab:L96-10D-Results} summarizes the performance of the three filters for different ensemble sizes averaged over 10 different simulations. It also shows the optimal bandwidths for the conditional diffusion model-based filters and the range (over all assimilation steps) of the number of adaptive steps taken to integrate \Cref{eq:reverse}.  From this table, we observe that the diffusion filter outperforms the EnKF and SIR filters for small to moderate ensemble sizes ($N\leq250$). This is also evident from \Cref{fig:L96-10D}, which visualizes the performance of the three filters over a window of 100 assimilation steps on one of the simulations for ensemble size $N=100$. Specifically, \Cref{fig:L96-10D} shows the observations, the corresponding true values of the states, the estimated mean of the assimilated states, the absolute error between the true states and the estimated mean, and the corresponding spread (one standard deviation) around the mean. In each of these plots, the horizontal axis represents assimilation steps, and the vertical axis represents the degree-of-freedom (dof) index. From the plot of the true solution we can observe wave-like patterns that are the characteristics of the Lorenz-96 system.  

A visual comparison of the plots of the mean for the three methods with the true state demonstrates that the diffusion model better matches the wave-like patterns observed in the true state. Both the EnKF and the SIR results also contain wave-like results, however, the patterns do not match the true state. This discrepancy is evident in the absolute error plots, which show that the errors for the EnKF and SIR models are much larger. Finally, plots of the standard deviation (termed Spread) reveal that the EnKF and the SIR filters have much smaller variability, and given that they incur significant error, they are overly confident in their estimates.

\begin{figure}[t]
    \centering
    \includegraphics[width=1.0\textwidth]{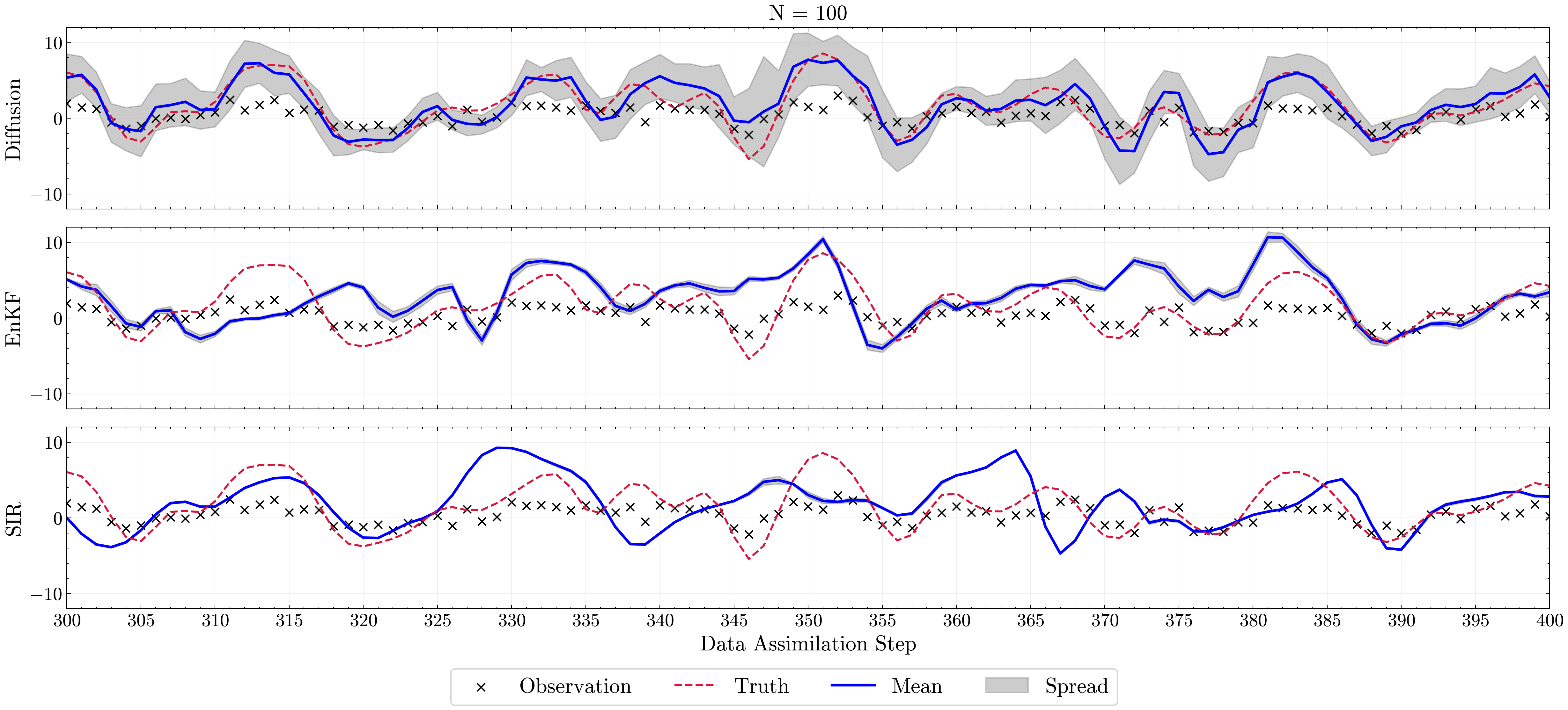}
    \caption{Performance comparison of three filtering methods for a single degree of freedom, $x_1$, of the 10-dimensional Lorenz-96 system with ensemble size $N=100$ over data assimilation steps 300-400, using the same test trajectory as in \Cref{fig:L96-10D}. The top, middle, and bottom plots correspond to the conditional diffusion model-based filter, EnKF, and SIR filter, respectively. In each plot, the vertical axis corresponds to the value of $x_1$, and the horizontal axis corresponds to the data assimilation steps. The solid blue line represents the ensemble mean of the assimilated state, the dashed red line denotes the true state trajectory, the black crosses indicate the observations, and the shaded gray region shows the ensemble spread (one standard deviation about the estimated mean).}
    \label{fig:L96-10D_timeplot}
\end{figure}

These observations are further confirmed in \Cref{fig:L96-10D_timeplot}, which plots a single dof, $x_1$, as a function of the assimilation step. We observe that the estimated mean (represented by the blue line) using the proposed approach aligns closer to the true trajectory (shown as the dashed red line) when compared with the other methods. \Cref{fig:L96-10D_timeplot} also illustrates that, unlike other approaches, the true state usually lies within the spread estimated using the diffusion filter.

From \Cref{tab:L96-10D-Results} we also note that the EnKF filter is more accurate than the diffusion filter for large ensemble sizes ($N\geq500$). This is expected because the filtering distribution is unimodal, and previous studies have shown that for the EnKF the estimated mean of the states closely follows the true states despite the nonlinearity in the system and the observation model~\cite{bao2024ensemble}. We would also like to remark that the fact that a method generates a mean that this closer to the true state does not imply that it generates a accurate representation of the true distribution. In order to test the closeness of the estimated distribution with the true distribution, we would need to generate samples from a benchmark distribution using a method like the SIR with a very large number of particles. However, this is computationally prohibitive for system of this size.

\subsubsection{Lorenz-96 in 20-dimensions}

\begin{table}[t]
\centering
\caption{Optimal kernel bandwidths $(\sigma_x, \sigma_y)$, the range of sampling steps used by the adaptive numerical integration scheme, and the corresponding time-averaged RMSE error $\mathcal{E}_{RMSE}$ of the conditional diffusion model-based filter with varying ensemble sizes $N$ for the 20-dimensional Lorenz–96 system. The time-averaged RMSE errors for the EnKF and SIR filters are also reported. All errors are also averaged over $S=10$ independent simulations. We indicate the best performing filter using bold fonts.}
\label{tab:L96-20D-Results}
\begin{tabular}{ccccccc}
\toprule
\multirow{2}{*}{\makecell[c]{\textbf{Ensemble size}\\N}} & \multicolumn{2}{c}{\textbf{Optimal Bandwidth}} &  \multirow{2}{*}{\makecell[c]{\textbf{Range of}\\\textbf{sampling steps}}}&\multicolumn{3}{c}{\textbf{Average $\mathcal{E}_{RMSE}$}} \\
\cmidrule(rl){2-3}
\cmidrule(rl){5-7}
& {$\sigma_x$} & {$\sigma_y$} & & {Diffusion} & {EnKF} & {SIR} \\
\midrule
20   & 0.20  & 1.50  & 13--17 & \textbf{3.550} & 5.251 & 5.008 \\
50   & 0.15 & 1.00  & 15--17 & \textbf{2.904} & 5.071 & 4.882 \\
100  & 0.15 & 0.75 & 15--17 & \textbf{2.456} & 4.842 & 5.000 \\
250  & 0.10  & 0.75 & 16--19 & \textbf{2.074} & 3.910 & 4.978 \\
500  & 0.10 & 0.75 & 17--19 & \textbf{1.771} & 2.196 & 4.948 \\
1000 & 0.10  & 0.50  & 17--19 & 1.439 & \textbf{0.747} & 4.902 \\
\bottomrule
\end{tabular}
\end{table}
\begin{figure}[!t]
    \centering
    \includegraphics[width=1.0\textwidth]{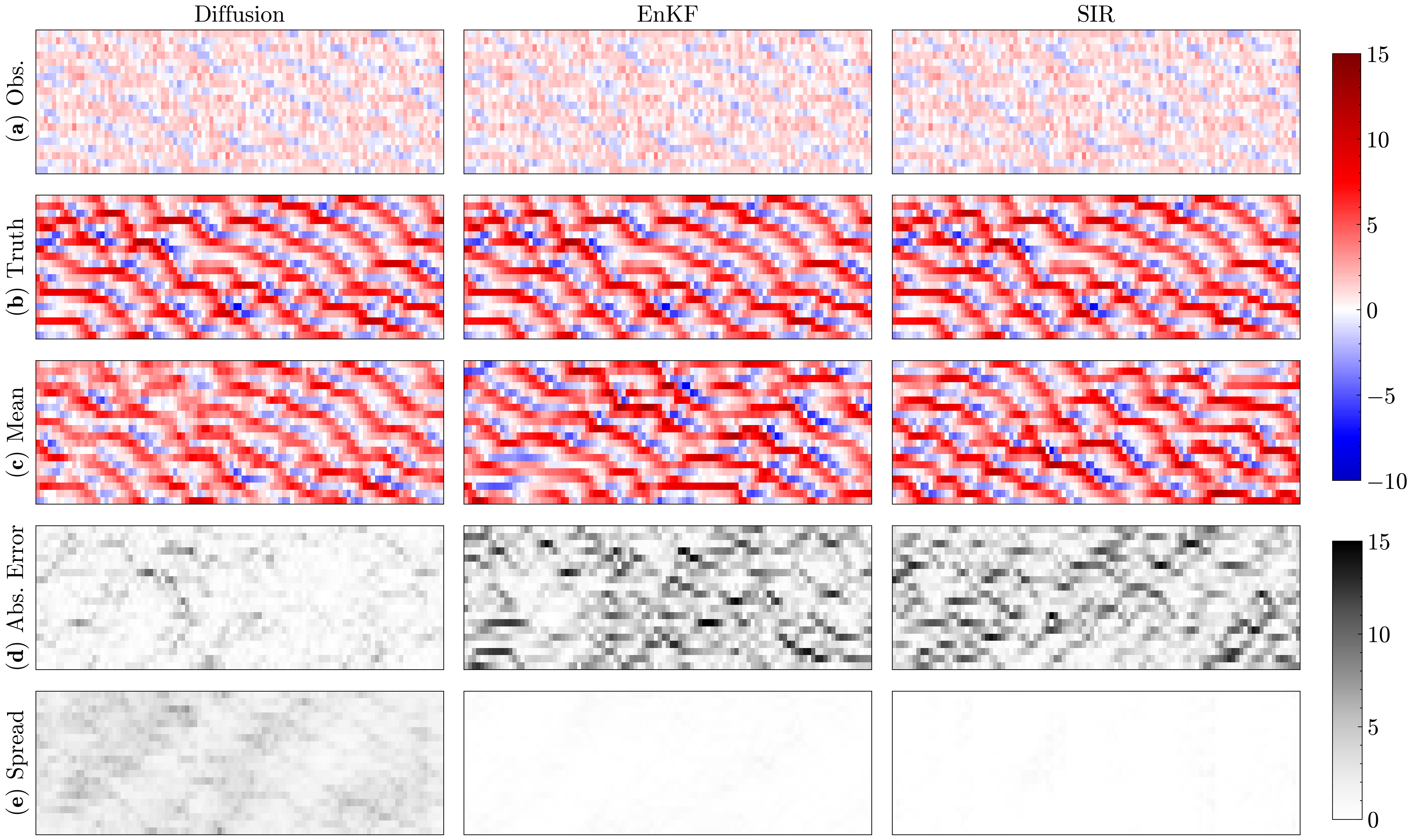}
    \caption{Visualization the performance of different filters with ensemble size $N=100$ for one test trajectory of the 20-dimensional Lorenz-96 system over data assimilation steps 300-400. In each plot, the vertical axis corresponds to the states, and the horizontal axis corresponds to the data assimilation step. Rows 1 through 5 correspond to the observations, true states,  estimated mean of the assimilated states, absolute error of mean with respect to the ground truth, and estimated ensemble spread (one standard deviation about the estimated mean), respectively. Columns 1 through 3 correspond to the conditional diffusion model-based, EnKF, and SIR filters, respectively.}
    \label{fig:L96-20D}
\end{figure}

Next, we consider the Lorenz-96 system in 20 dimensions. \Cref{tab:L96-20D-Results} summarizes the performance of the three filters for different ensemble sizes $N$ using $\mathcal{E}_{RMSE}$ averaged over 10 different simulations. It also shows the range of the number of adaptive steps taken to integrate \Cref{eq:reverse}. One important remark we make from \Cref{tab:L63-Results,tab:L96-10D-Results,tab:L96-20D-Results} is that the number of steps necessary to integrate \Cref{eq:reverse} does not increase with increasing dimensionality $d$ of the problem. Additionally, \Cref{tab:L96-20D-Results} again shows that the diffusion filter outperforms the EnKF and SIR filters for small to moderate ensemble sizes ($N\leq 500$).
\begin{figure}[!t]
    \centering
    \includegraphics[width=1.0\textwidth]{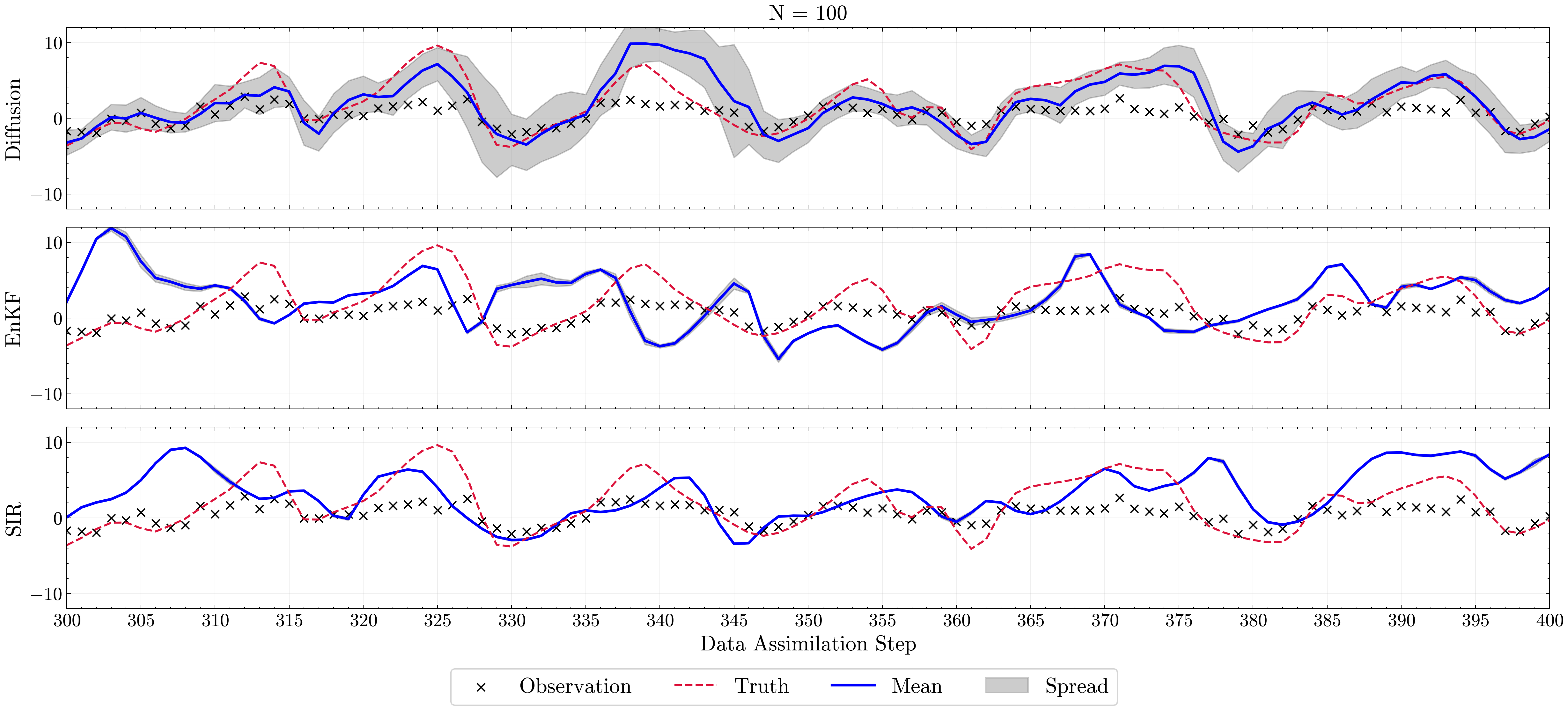}
    \caption{Performance comparison of three filtering methods for a single degree of freedom, $x_1$, of the 20-dimensional Lorenz-96 system with ensemble size $N=100$ over data assimilation steps 300-400, using the same test trajectory as in \Cref{fig:L96-20D}. The top, middle, and bottom plots correspond to the conditional diffusion model-based filter, EnKF, and SIR filter, respectively. In each plot, the vertical axis corresponds to the value of $x_1$, and the horizontal axis corresponds to the data assimilation steps. The solid blue line represents the ensemble mean of the assimilated state, the dashed red line denotes the true state trajectory, the black crosses indicate the observations, and the shaded gray region shows the ensemble spread (one standard deviation about the estimated mean).}
    \label{fig:L96-20D_timeplot}
\end{figure}

In \Cref{fig:L96-20D}, we plot results from the true state and the three filters over a window of 100 assimilation steps for one of the test cases with an ensemble size $N=100$. In this figure, we have plotted the observations, the corresponding true value of the states, the estimated mean of the assimilated states, the absolute error between the true states and the estimated mean, and the corresponding estimated spread (one standard deviation). Once again, in each plot  the horizontal axis represents assimilation steps and the vertical axis represents the dof index. 

Similar to the $d=10$ case, we observe wave-like solutions in the true state plot. The patterns are also observed in the plots for the estimated mean for each method, where the diffusion model appears to be most accurate. This is borne out by the plot of the absolute error, which have much larger values for the EnKF and SIR filters. Further, as in the $d=10$ case, the EnKF and SIR filters yield very small standard deviations (seen in the Spread plots), indicating they are inaccurate and overly confident in their predictions. This behavior is further illustrated in \Cref{fig:L96-20D_timeplot}, which presents the results for a single dof, $x_1$. As in the $d=10$ case, the ensemble mean obtained from the proposed diffusion-based approach (shown in blue) more closely tracks the true trajectory (dashed red) and the corresponding spread more consistently encompasses the true trajectory than the other filters. 

\section{Conclusion}\label{sec:conclusion}

In this work, we have explored closed-form conditional diffusion models for data assimilation. Specifically, we employ these  models to solve the inverse problem arising in the update step of the Bayes filter. This formulation requires paired samples of the state and the corresponding measurements, which can be computed by applying the process and observation models as black-boxes to samples from the previous assimilation step. 

Starting from a kernel density estimate of the joint empirical distribution of the state and its corresponding measurements, we show that the score function can be evaluated analytically along the reverse-time process of the conditional diffusion model. This, in turn, enables efficient sampling from the filtering distribution via numerical integration. Because the proposed approach is entirely sample-based, it can readily accommodate process and measurement models without requiring explicit knowledge of their underlying distributional forms.

We also investigate the performance of the proposed closed-form conditional diffusion models on nonlinear data assimilation problems involving the Lorenz–63 and Lorenz–96 systems. The numerical examples include systems that are chaotic, of moderate dimensionality (up to 20), and involve observation models that are nonlinear or induce strongly non-Gaussian and bimodal filtering distributions. Compared to widely used filters, such as the EnKF and the SIR filter, the results demonstrate that the proposed approach performs well in approximating complex non-Gaussian filtering distributions, particularly when the ensemble size is small to moderate. We note that this ability to produce accurate results with a small ensemble size is particularly valuable in systems where the forward model is highly complex and computationally expensive, such as those used to simulate weather or the spread of wildfires.

Future work will explore adaptive strategies for selecting the bandwidth parameters $\sigma_x$ and $\sigma_y$, as well as fast multipole methods to improve the computational efficiency of the proposed approach. Extensions to more practical applications, along with theoretical results establishing the relationship between problem dimensionality and the required ensemble size, also represent promising directions for future research.

\backmatter

\bmhead{Acknowledgments}
The authors acknowledge the Center for Advanced Research Computing (CARC, \href{https://carc.usc.edu}{carc.usc.edu}) at the University of Southern California for providing computing resources that have contributed to the research results reported within this publication. 

\section*{Declarations}

\bmhead{Funding} 
This work was initiated while AD was a postdoctoral researcher at the University of Southern California. The authors acknowledge support from ARO grant W911NF2410401, ARO cooperative agreement W911NF-25-2-0183 and NSF Award \# 2231659. AD was additionally supported by the John von Neumann Fellowship at Sandia National Laboratories. 

Sandia National Laboratories is a multi-mission laboratory managed and operated by National Technology \& Engineering Solutions of Sandia, LLC (NTESS), a wholly owned subsidiary of Honeywell International Inc., for the U.S. Department of Energy’s National Nuclear Security Administration (DOE/NNSA) under contract DE-NA0003525. This written work is authored by an employee of NTESS. SAND2026-19194O. The employee, not NTESS, owns the right, title and interest in and to the written work and is responsible for its contents. Any subjective views or opinions that might be expressed in the written work do not necessarily represent the views of the U.S. Government. The publisher acknowledges that the U.S. Government retains a non-exclusive, paid-up, irrevocable, world-wide license to publish or reproduce the published form of this written work or allow others to do so, for U.S. Government purposes. The DOE will provide public access to results of federally sponsored research in accordance with the DOE Public Access Plan \url{https://www.energy.gov/downloads/doe-public-access-plan}.

\bmhead{Conflict of interest/Competing interests}
The authors declare no conflicts of interest.

\bmhead{Ethics approval and consent to participate}
Not applicable.

\bmhead{Consent for publication}
All authors provide consent to publish.

\bmhead{Data availability}
Data will be made available upon reasonable request.

\bmhead{Materials availability}
Not applicable.

\bmhead{Code availability}
Code will be made available upon reasonable request.

\bmhead{Author contribution}
B.B. conceptualized the research, developed the code, designed and implemented the algorithm, conducted the simulations, analyzed the results, made the visualizations, and drafted the initial manuscript.  A.D conceptualized the research, developed the code, and drafted the initial manuscript. A.O. conceptualized the research, acquired funding, provided supervision and resources, and drafted the initial manuscript. All authors reviewed and approved the final version of the manuscript.

\begin{appendices}

\section{Gaussian Kernel Properties}\label{appsubsec:Gaussian-Kernel-Properties}
\subsection{Convolution of two Gaussian Kernels}\label{appsubsec:convolution-Gaussians}
First, we define the Fourier Transform and its associated Inverse Fourier Transform as follows:
\begin{equation}\label{eq:forward_fourier}
    \mathcal{F}(f(\x))(\bm{k}) 
    = \int_{\mathbb{R}^d}f(\x) e^{-2\pi i \x \cdot \bm{k}} d\x ,
\end{equation}
\begin{equation}\label{eq:inverse_fourier}
    \mathcal{F}^{-1}(F(\bm{k}))(\x)
    = \int_{\mathbb{R}^d}F(\bm{k}) e^{2\pi i \x \cdot \bm{k}} d\bm{k} .
\end{equation}
Now, consider the convolution theorem which states that the Fourier transform of a convolution of two functions is the product of their Fourier transforms. Thus,
\begin{equation}\label{eq:convolution_theorem}
    f(\x) \ast h(\x) 
    = \int_{\mathbb{R}^d} f(\x - \x') h(\x') d\x'
    = \mathcal{F}^{-1}(\mathcal{F}(f(\x))\mathcal{F}(h(\x))).
\end{equation}
Also note some useful properties of the Dirac Delta function:
\begin{eqnarray}
    \delta(\x) &=& \delta(-\x), \\
    \delta(\x) &=& \frac{1}{(2\pi)^d} \int_{\mathbb{R}^d} e^{i\bm{k} \cdot \x} d\bm{k} 
    = \frac{1}{(2\pi)^d} \int_{\mathbb{R}^d} e^{i(2\pi \bm{\xi}) \cdot \x} (2\pi)^n d\bm{\xi} 
    = \int_{\mathbb{R}^d} e^{2\pi i \bm{\xi} \cdot \x} d\bm{\xi},
\end{eqnarray}

Recall, we defined a Gaussian kernel in \Cref{eq:gaussian_kernel}. We can evaluate what the Fourier transform of an arbitrary multivariate, isotropic Gaussian kernel as follows:
\begin{eqnarray}
    \mathcal{F}(g_\sigma(\x-\bm{\mu})) (\bm{k})
    &=& \int_{\mathbb{R}^d} g_\sigma(\x-\bm{\mu}) e^{-2\pi i \x \cdot \bm{k}} d\x \nonumber\\
    &=& \frac{1}{(2\pi \sigma^2)^{d/2}} \int_{\mathbb{R}^d} 
    \exp \left( -\frac{\lVert \x - \bm{\mu} \rVert^2_2}{2\sigma^2} \right)
    e^{-2\pi i \x \cdot \bm{k}} 
    d\x \nonumber\\
    &=& \frac{e^{-2\pi i \bm{\mu} \cdot \bm{k}}}{(2\pi \sigma^2)^{d/2}} \int_{\mathbb{R}^d} 
    \exp \left( -\frac{\lVert \x - \bm{\mu} \rVert^2_2}{2\sigma^2} \right)
    \exp(-2\pi i (\x-\bm{\mu}) \cdot \bm{k})
    d\x \nonumber\\
    &=& \frac{e^{-2\pi i \bm{\mu} \cdot \bm{k}}}{(2\pi \sigma^2)^{d/2}} \int_{\mathbb{R}^d} 
    \exp \left[ 
    -\frac{1}{2\sigma^2} \left(
    \lVert \x - \bm{\mu} \rVert^2_2 
    + 2(2\pi\sigma^2 i (\x-\bm{\mu}) \cdot \bm{k})
    \right)\right]
    d\x \nonumber\\
    &=& \frac{e^{-2\pi i \bm{\mu} \cdot \bm{k}}e^{\frac{1}{2\sigma^2} \lVert 2\pi i \sigma^2 \bm{k} \rVert^2_2}}{(2\pi \sigma^2)^{d/2}} \int_{\mathbb{R}^d} 
    \exp \left[ 
    -\frac{1}{2\sigma^2} \lVert \x - \bm{\mu} + 2\pi i\sigma^2 \bm{k}\rVert^2_2 \right]
    d\x \nonumber\\
    &=& \frac{e^{-2\pi i \bm{\mu} \cdot \bm{k}}e^{-2\pi^2\sigma^2 \lVert \bm{k} \rVert^2_2}}{(2\pi \sigma^2)^{d/2}}
    (2\pi\sigma^2)^{d/2} \nonumber\\
    &=& \exp(-2\pi i \bm{\mu} \cdot \bm{k})\exp(-2\pi^2\sigma^2 \lVert \bm{k} \rVert^2_2)
\end{eqnarray}
By definition of Fourier Transforms:
\begin{equation}
    \mathcal{F}^{-1}(\mathcal{F}(g_\sigma(\x-\bm{\mu}))
    = \mathcal{F}^{-1}(\exp(-2\pi i \bm{\mu} \cdot \bm{k})\exp(-2\pi^2\sigma^2 \lVert \bm{k} \rVert^2_2))
    =g_\sigma(\x-\bm{\mu})
\end{equation}
Therefore, the convolution of two Gaussian kernels:
\begin{eqnarray}
    g_{\sigma_f}(\x-\bm{\mu}_f) \ast g_{\sigma_h}(\x-\bm{\mu}_h) 
    &=& \mathcal{F}^{-1}(\mathcal{F}(g_{\sigma_f}(\x- \bm{\mu}_f))\mathcal{F}(g_{\sigma_h}(\x-\bm{\mu}_h))) \nonumber\\
    &=& \mathcal{F}^{-1}(
    e^{-2\pi i \bm{\mu}_f \cdot \bm{k}}e^{-2\pi^2\sigma_f^2 \lVert \bm{k} \rVert^2_2}
    e^{-2\pi i \bm{\mu}_h \cdot \bm{k}}e^{-2\pi^2\sigma_h^2 \lVert \bm{k} \rVert^2_2}) \nonumber\\
    &=&  \mathcal{F}^{-1}(
    e^{-2\pi i (\bm{\mu}_f+\bm{\mu}_h) \cdot \bm{k}}
    e^{-2\pi^2(\sigma_f^2+\sigma_h^2) \lVert \bm{k} \rVert^2_2}) \nonumber\\
    &=& g_{\sqrt{\sigma_f^2+\sigma_h^2}}(\x-(\bm{\mu}_f+\bm{\mu}_h))
\end{eqnarray}
is also a Gaussian Kernel. 

\subsection{Gradient of a Gaussian Kernel}\label{appsubsec:gradient_Gaussian}

\begin{eqnarray}
    \nabla_{\bm{x}} g_{\sigma}(\x) 
    &=& \nabla_{\bm{x}} \left[\frac{1}{(2\pi \sigma^2)^{d/2}} \exp \left(-\frac{\lVert \x \rVert^2_2}{2 \sigma^2}\right)\right] \nonumber\\
    &=& \frac{1}{(2\pi \sigma^2)^{d/2}} \nabla_{\bm{x}} \exp \left(-\frac{\lVert \x \rVert^2_2}{2 \sigma^2}\right) \nonumber\\
    &=& \frac{1}{(2\pi \sigma^2)^{d/2}}  \exp \left(-\frac{\lVert \x \rVert^2_2}{2 \sigma^2}\right) \nabla_{\bm{x}} \left(-\frac{\lVert \x \rVert^2_2}{2 \sigma^2}\right) \nonumber\\
    &=& g_{\sigma}(\x) \left(-\frac{\x}{\sigma^2}\right)
\end{eqnarray}

\end{appendices}

\bibliography{references}

\end{document}